# USING ONTOLOGY-BASED CONTEXT IN THE PORTUGUESE-ENGLISH TRANSLATION OF HOMOGRAPHS IN TEXTUAL DIALOGUES

Diego Moussallem and Ricardo Choren

Military Institute of Engineering, Dept. Computer Engineering,
Pr. General Tiburcio - Rio de Janeiro, 22290-270

## ABSTRACT

*This paper introduces a novel approach to tackle the existing gap on message translations in dialogue systems. Currently, submitted messages to the dialogue systems are considered as isolated sentences. Thus, missing context information impede the disambiguation of homographs words in ambiguous sentences. Our approach solves this disambiguation problem by using concepts over existing ontologies.*

## 1. INTRODUCTION

The ability of people to communicate each other becomes increasingly important due to the conducting of trade agreements, implementation of academic activities, or just aiming the intention of a relationship [1]. There are several forms of communication, the most important is the *dialogue*. The dialogue starts whenever someone who wants to start it searches for appropriate words in order to fashion an understandable message to another participant. Initially, each statement requires an answer or a new statement counterpart [2]. With technological growth, dialogue systems have become popular.

Generally, a dialogue system aims to provide support for the communication (exchange of messages) among people using a single language. When the players use different languages, the dialogue system needs to perform the translation of the messages. In order to achieve that, several translation systems have been developed, such as *Bing Translator*, *Google Translator*, *Systran*, *Worldlingo*, *and Gengo*. However, during a dialogue, homographs terms can occur, i.e. words which have the same spelling but different meanings. These words may generate ambiguity at the time of translation because, according to *Bar-Hillel* [3], the flow for selecting the meaning of given homograph word in an automated translation tool relies on a statistical algorithm. Translating homographs in sentences into dialogue is a challenging task: most translation techniques, such as Transfer-based, Example-based, Statistical are developed for retrieving the translation based on frequency algorithms [4]. These algorithms are designed to decide through pre-established criteria, which is the better translation. This process leads partially to solve the problem of homographs though.

For example, when sending the sentence "the bank has a problem" with the homograph "bank", the translators result  as "the bank is in trouble" more likely because the translation of "bank" is the one with the highest frequency in the database. However, if we were talking about another kind of bank, as a "seat" or a "database", the translation would result in error. To solve this problem is possible to think that a dialogue is interaction through a language in that all share same context [5]. In fact, the ambiguity of words homographs is directly connected at the context in which a word is ambiguous when it has different meanings depending on the subject.

The translation tools are able to find what is the context being used. However, to define the context of a text, it must state extensiveness [6], so the translator can identify into the text, which is the correct meaning for the ambiguous term based on neighbouring words.

The existing works about word sense disambiguation try to solve the ambiguity problem in text, websites, and applications [7]. There are several solutions that achieving good results, but the focus of this work is apply the disambiguation in dialogue systems. Therefore, Works like *Carput and Wu* [8], *Chiang et. al.* [9] and others that are quite important to word sense disambiguation do not apply in this word. Because, when the user send the short sentence in dialogue and it use a machine translation is necessary to be defined at least some pre-context. In section of related works is presented two papers that try to disambiguate the ambiguous words, however *Gutierrez et. al.* [10] works uses statistical methods after semantic step it is not so good for disambiguate short sentences. In *Harriehausen-Muhlbauer and Heuss* [11] an interesting proposal but your performance is very bad to use in a dialogue system and the authors present simple examples but the approach is very interesting to the machine translation future.

This work proposes an approach for homograph translation disambiguation that relies on dialog context obtained through ontologies. The focus is to assist on the translation of messages in communication tools, i.e. for short text snippets. This method performs a pre-translation of homographs words using the ontologies for understanding of the context and with that machine translation gets to provide a more accurate translation into a dialogue.

Specifically, this work makes the following contributions: A method for disambiguation of homographs terms in multilingual dialogue systems. Achieving a better accuracy of results in machine translation using sematic web technologies. This method has been developed just for the Portuguese language since each language present a specific grammatical syntax. The method is applied to a dialogue between the Portuguese and English languages.

The results suggest the approach has useful function to disambiguate the homographs in dialogue and also it got to show the relevance in using semantic web technologies as knowledge-base to machine translation when applied in a dialogue.

The structure of the paper is as follows: Section 2 describes the translation process with dialogue while Section 3 presents our novel method to decrease ambiguity and we present examples of application and evaluation of this method in Section 4. In Section 5 describes the state-of-the-art and discussed. We conclude and pointing out future works in Section 6.

## 2. BACKGROUND AND MOTIVATING EXAMPLES

In the dialogue or any other way of communication, the needing of defining the context exists, so it can be properly materialized. The context has a high important for the understanding of the words and the consent of the meaning of each term present in a message [12]. For example, in the sentence in Portuguese Language "A bateria esta com defeito", there are two types of battery when translating to English. Battery as ``drums" (musical instrument) or ``battery" (electronic component).

A further example is "A vela está ruim" with three different contexts: sail boat (sail) and car (spak plugs) or household candle (candle). For each type of "candle", there is a translation in English. These phenomena occur in all languages, such as English language the word "pipe" in the sentence "I need to buy a pipe". You need to buy a "cachimbo" or a "tubo", both in the English language are written the same way, but when we translate into Portuguese is described in more than one word depending on the context [13].

Currently translations in a dialogue are carried out using independent machine translations. The most of machines use statistics model, thus there is a possibility that it hits (to a dialogue context) the translation of homograph word [13] [14]. Following table presents results about the translation cited before, thus confirming the use of statistics models to small sentences.

Sentence 1 "O banco está com problema."
Sentence 2 "A bateria está com defeito."
Sentence 3 "A vela está ruim."

Table 1. Experiments with Tools: What is mean of the homograph word?.

| Systems | Sentence 1 | Sentence 2 | Sentence 3 |
| --- | --- | --- | --- |
| Bing Translator | Bank | Battery | Candle |
| Google Translator | Bank | Battery | Candle |
| Worlingo | Bank | Battery | Candle |
| Systran | Bank | Battery | Candle |
| Gengo | Bank | Battery | Candle |

To solve this problem, a choosing would be about to perform a semantic analysis of messages sent in a dialogue. Semantic analysis is responsible for defining the meaning of the words [15]. It is possible to think that the right meaning of the word is directly connected the conceptual issue of words. Thus, Bag of words could be used as a solution to the problem of homographs translation.

However, a bag of words can be used of two ways: First, providing power of choice to user or using statistical models based on words frequency. When the user chooses the meaning, that answer is stored and then continue using the statistical as sort way. Therefore, this solution is not suitable to dialogue system. [16]

Context in semantic web area is handled through ontologies [17]. Ontologies provide concepts, meanings about a specific domain [18]. For more complex that is an ontology repository can be extract specific concepts, the meaning of specific word and a relationship among entities.

The use of ontologies to selecting the right meaning of homograph word in a dialogue is equivalent to the conceptual and philosophic question into in ontologies. In according *Guarino* [18], the ontologies describe real or literary concepts about things.

When using an ontology to support in translation of an ambiguous term, it is not only exchanged to word by another. This exchange is loaded all the conceptual question of word, its origin and its meaning about subject discussed in the dialogue. This way, the use of ontologies to support in translation task of homographs words shows promising, since it relies on dialogue context to determine its meaning.

## 3. METHOD JUDGE

This section presents comprehensively the development of the novel method. This method aims to support the automatic translators to select a translation more suitable to homographs terms in dialogue systems. Figure 1 represent an optical about overview process of JUDGE Method.

First step called *Semantic Verification of Homographs Words* is responsible for choose the meaning of ambiguous word. This step is split in two activities: *Morphological Analysis*, and; *Semantic Analysis*. Second step *Translation* is responsible for realize the translation of message to be sent through dialogue system. This step also is split in two activities: *Automatic Replacement*, and; *Automatic Translation*.

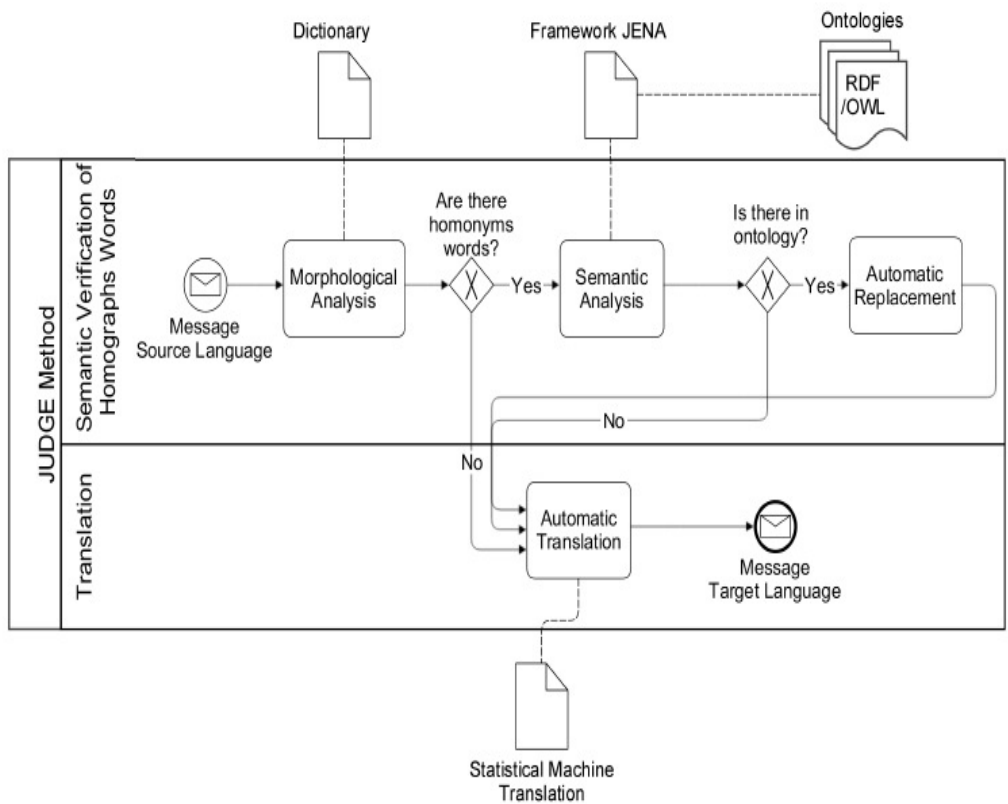

Figure 1. Overview Method

In figure 2 is presented the method step by step when the method receive the user message.

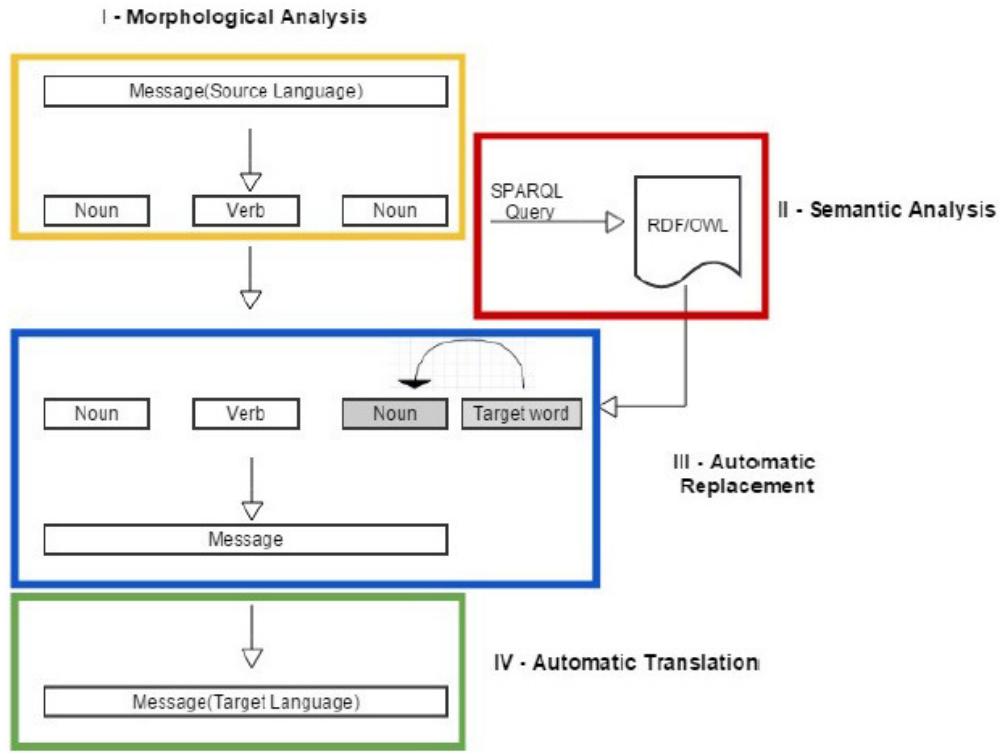

Figure 2. Method Steps

## 3.1. Morphological Analysis

Morphological analysis is responsible for fragmenting the sentence to be done the search of words in dictionary. Dictionary consists in a data repository containing grammatical information about each word of specific language. According to *Cegalla* [19], the most of homographs words are nouns or verbs. With that was defined as requisition that each word that owned this part of speech like noun or verb must be looked in ontology.

Dictionary is used in morphological analysis as label way the part of speech of each word contained in sentence sent by user. Morphological analysis is seen as a rule in all work about disambiguation words. With it is possible to identify not only part of speech, but also words origin, its syntax, among others possibility of sub-analysis. Therefore, the use of data dictionary is vital to the method.

Due to method is developed to Portuguese language, it was necessary to give more attention to phenomenon of locutions. This phenomenon is seen as a variation of language which implies directly in homographs words. Because, in some cases the homographs used locutions to describe it. According with *Rocha Lima* [20] a substantive or adjective locution are formed to a preposition added of a noun. For instance, the word "banco" has several meanings like financial institution, a database or seat. However this word is used by many time as "banco da varanda", "banco de dados", among others. The same happens with others homographs words like "bateria" being "bateria do carro" or "bateria da banda". Therefore in this variation the homographs words have its meanings defined by its word subsequently.

With that was necessary to develop an algorithm in method to do addressing of locutions. In case the subsequent words have its parts of speech like preposition and noun, the method does not query the word meaning in the ontology. Thus, the statistical machine translation assumes the responsibility to solve the ambiguity making the translations based in neighbour's words. With that machine translation will be able to identify the context based on its complement. Thus finding the right meaning of word. [21]

A sub-problem found in morphological analysis was the part of speech of word "da" in locutions. For instance, the sentence "a bateria da banda não chego". Part of speech attribute by dictionary to "da" was "verb" having its origin in verb "da" (give). The right meaning would be "preposition" to be a word responsible for connect two words. To solve this problem was created a detection about this word in sentences.

By definition a sentence has only one verb [22]. Therefore, when there is a verb and the word "da" in same sentence, the method changes the part of speech to preposition. With that judging its class of right form in according with the sentence. When there is not other word "da" in same sentence and it is subsequent of noun, its class is considered as verb being original result of dictionary.

To be found the part of speech of each term and the term is a possible candidate it submitted to next activity.

### 3.2. Semantic Analysis

After morphological analysis activity, the homographs candidates are separated. These candidates are queried in the ontologies to obtain the right meaning. Ontology is responsible for getting right meaning and return the writing of the word in target language. Ontology is used in semantic analysis activity to define the right context to ambiguous term in dialogue.

The ontology is selected based on log messages of user. The selection occur when the ambiguous word is found log messages. The word is queried in matrix correlation created with based on word frequency in messages to each context. The context is found due to occurrence of specific words in corpus message using annotation skill. E.g. the words guitar, band, concert and lyrics to music domain. In Table 2 is presented a part of matrix correlation.

Table 2. Matrix Correlation.

| Words | Context | | | | |
|---|---|---|---|---|---|
| | Music | Electronic/Computer | Vehicles | Sports | Financial |
| Bateria | 78% | 40% | 32% | 2% | 0% |
| Vela | 0% | 2% | 63% | 92% | 0% |
| Banco | 0% | 80% | 56% | 0% | 12% |
| Bolsa | 0% | 0% | 0% | 11% | 73% |
| Rede | 0% | 92% | 0% | 0% | 0% |

After the choosing of ontology, the search in ontology is divided in two tasks. First task performs a scan to ensure the occurrence of term in ontology. This is necessary to ensure the simultaneous characteristics, because it would has a waste of performance seeking a term there is not in ontology. Therefore being confirmed the existence of term, to start second task. That is realized a search in *SPARQL* [23] on ontology to retrieve the translation of ambiguous word queried.

For instance, "a bateria está quebrada" (Portuguese), the semantic analysis seeks in ontology to find the true meaning of word "bateria". Ontology is needed to make the disambiguation. Ontology used in this example was *Music Ontology* [24], it is responsible for describing the music domain, from musicians to musical instruments. That ontology uses the *SKOS* pattern to depict its properties. *SKOS* has as purpose to describe vocabularies about specific domain and label it in ontologies [25]. Those vocabularies may be described in more than one language. That label form of *SKOS* is crucial in evolution of the method and it seen as the future to machine translation using semantic web technologies.

However, Music Ontology description was only in English language as the most of ontologies [26]. With that needed to add information in Portuguese language to be possible to use it in method. Thus was possible to apply it to solve the ambiguity of term "bateria", in this case referring as musical instrument.

```
Music Ontolgy Original

<skos:narrower>
    <skos:Concept rdf:about="http://purl.org/ontology/mo/mit#Drums">
     <skos:prefLabel xml:lang="en">Drums</skos:prefLabel>
      <skos:inScheme rdf:resource="http://purl.org/ontology/mo/instruments#Musical_instruments"/>
     <skos:narrower>

Music Ontology Modify

<skos:narrower>
    <skos:Concept rdf:about="http://purl.org/ontology/mo/mit#Drums">
     <skos:prefLabel xml:lang="en">Drums</skos:prefLabel>
     <skos:prefLabel xml:lang="pt">Bateria</skos:prefLabel>
      <skos:inScheme rdf:resource="http://purl.org/ontology/mo/instruments#Musical_instruments"/>
     <skos:narrower>
```

Figure 3. Music Ontology excerpt

In the figure 3 is presented before and after of ontology. Data inserted can be seen in *Music Ontology Modify*. It was added the word "bateria" describing that the term "Drums" in Portuguese is written this way. After this modification was possible the execution of semantic activity in method. The scan was made with success indicating that the term there is in ontology. This scan is done with owner mechanism of component semantic, in this case *JENA* [27]. With the confirmation is started the query in *SPARQL*.

Figure 4 features a query in *SPARQL* on music ontology. It was used *URI* of *SKOS* as prefix and namespace "skos" to perform the search. In the first part of code is inserted the subject of the triple. This subject was recovered through the scan performed in ontology. Second part is the property that need to be found. Third part is the value to be retrieved by ontology. In the last part of code is defined a filter to the result to be returned in English language. The result emitted from ontology was "drums" to this query in *SPARQL*.

```
"PREFIX skos: <http://www.w3.org/2004/02/skos/core#>

SELECT ?prefLabel
WHERE {<http://purl.org/ontology/mo/mit#Drums> skos:prefLabel ?prefLabel.
        FILTER (lang(?prefLabel) = "@en ")}";
```

Figure 4. Sparql Query

During the semantic analysis activity had been verified the necessity to preserve of context in dialogue. Preservation was needed because for more than one subject is defined, always occur the iteration of sub-subjects into same chat. The problem is appear sub-subjects using same homograph word in chat. When the homograph word appeared on next iteration in most of cases, it is directly linked to meaning of the preceding sentence.

Following an example of dialogue coordinating mismatch of contexts. Context used was vehicles. Following dialogue.

1ª iteration - I put the sparkplugs and the drums at the trunk. (Eu coloquei as velas e a bateria no porta malas.)
2ª iteration - ok, e a bateria estava completa? (Ok, and the battery was complete?)

In the above dialogue is possible to perceive that in first iteration the word ``bateria'' is identified as musical instrument "drums", even the context being about cars. When the second iteration is started referring the same ``bateria'', the translation result was "battery", it happened because the context established in dialogue was about cars indicated by the ontology. This mismatch occurred because when the messages are sent to machine translation it is not interconnected. It sends of independent way. Therefore, the method instead of get the right meaning, it retrieve the result provided by ontology and it replace the word in sentence.

To address this problem was used theory of dialogue. That served as basis to treat this problem. In according with *Brait* [2], a basic chat is achieved when a person send a message. Therefore, the receiver answer or he asks with another question the sender. Therefore, the dialogue can split in three fragments.

In according with theory, it was implemented in method through temporary log with a limit of three iterations. With that is stored the temporary meaning of words in sentences. When it is achieved the four iteration the temporary log is reset, thus returning the context indicated by ontology. With this treatment performed, when second iteration is sent instead of search the ontology, the method makes a query in temporary log. So is identified that the homograph word showed in previous sentence, thus is retrieved the right meaning. With dialogue theory applied in method the result of translation sentence was "ok, and the drums was complete?"

### 3.3. Translation

With the result ensured by ontology, the method starts automatic replacement technique. This technique was based on studies about statistical machine translation [28]. In statistical machine translation is possible ensuring that when is sent a word in source language among others words, in target language the same influence on translations of neighbours terms [29]. Automatic replacement activity is responsible for performing pre-translation of word. After the replacement, the sentence is again concatenated and submitted to statistical machine translation.

This activity abstained from necessity to develop a statistical machine translation, because there are many machine translation more robust that already implement this approach and they have a large corpus annotated. Therefore, it was used an independent component that it implements this approach. The way as the method was developed become it independent of the translation tools. Therewith it is possible to change the tool without compromising the functionality of method, since it implements the statistical approach.

In this activity, the sentence is submitted to machine translation to end the translation complete of sentence, thus becoming a translation more cohesive and meaningful to dialogue.

### 3.4. Other Languages

Method was idealized considering only the Portuguese language structure. Portuguese language is the sixth idiom most spoken in the world and also it has a huge population as natives speakers [30]. It was solved specific problems like verbal locution, words identified with more than one syntax function as word "da" can be verb or preposition.

For that, the method be suitable to other languages and universally used, there is necessity of grammatical development for each one them. Each language has its own grammatical order, there are similar cases such as Spanish, but it with the use of method without previous syntax treatment can generate many problems in translation. This approach cannot applied to idioms have declination in its words like German and Russian languages.

## 4. EXAMPLE OF APPLICATION

This section presents details about the development of dialogue system that implements the JUDGE method.

### 4.1. Tool

Tool was projected to be installed in some web server, since the same has support to *Java language*. System is composed for two administrative areas and one to dialogue. Administrative areas require authentication. With that authentication is realized managing user, making it possible the use of two or more user in each administrative area.

Before starting the dialogue system, it is still necessary to inform which is log message file to be uploaded to assignment the pre-context at dialogue. In dialogue, area is made the exchange of participant's messages. Messages use of the culture form to implementation examples. Source language is located into of parentheses beside target language as support to understanding and learning about languages.

### 4.2. Example – Isolated Sentence

This section illustrates the translation of sentence isolated using the method. In this example was translated a sentence "a minha bateria está ruim". It was used the ontology "Music Ontology" referent to music domain simulating a help desk about musical instrument. In figure 5 is presented an excerpt from chat. In Chat, the user Diego bought drums by the website and it was not working. Attendant Thomas supports Diego.

International Journal of Artificial Intelligence & Applications (IJAIA) Vol. 6, No. 5, September 2015

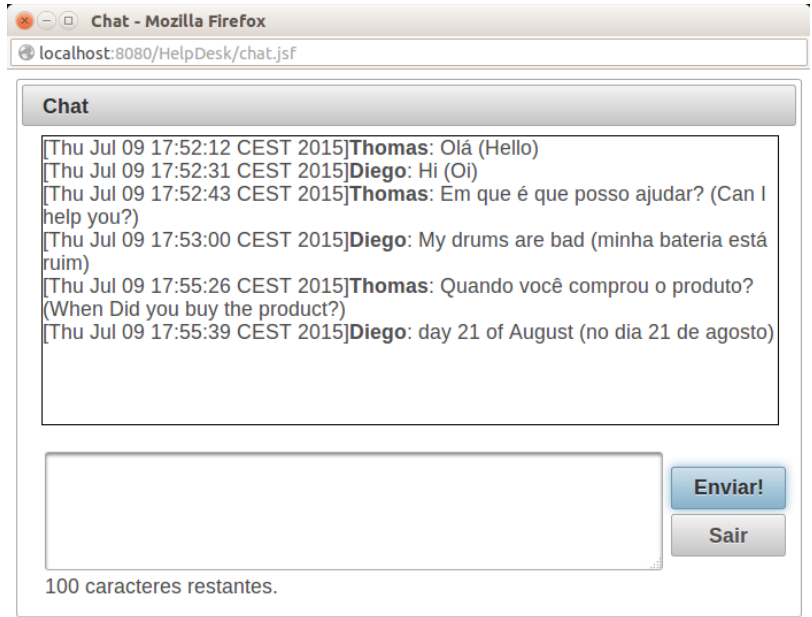

Figure 5. Dialogue - Music Context

In figure 6 is showed the method working based this example. Doing. When the message is sent, Analysis Morphological Activity is started realizing the fragmentation of terms contained in the sentence. These terms are queried in *OpenNLP* [31], which returns part of speech of each word. Word "minha" was classified as pronoun, "bateria" as a noun, "está" being verb and "ruim" as an adjective. The words "bateria" and "está" are submitted to semantic analysis in according with the method specification. These words are queried in *SPARQL* on ontology using *SKOS* pattern. This query has the goal of recover only the writing in another language, in case the English language.

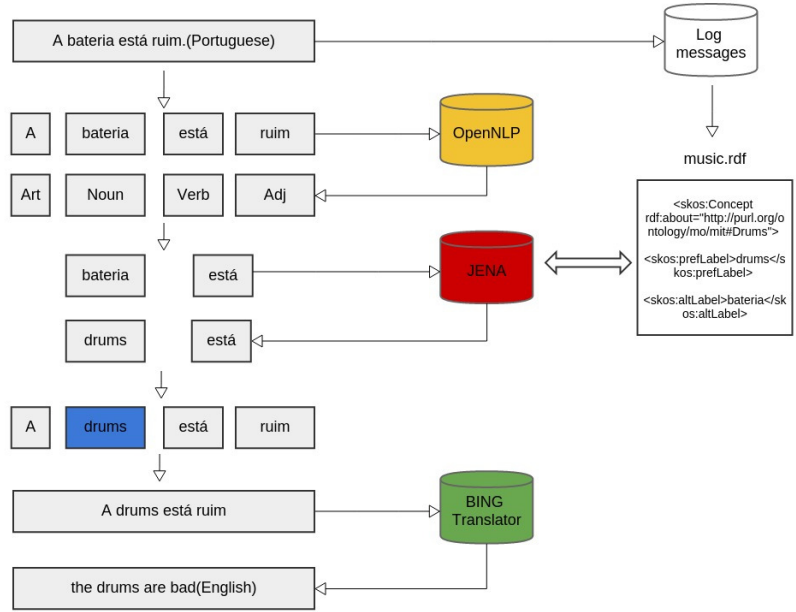

Figure 6. Example Isolated Sentence working.

International Journal of Artificial Intelligence & Applications (IJAIA) Vol. 6, No. 5, September 2015

Query received as return the word "drums", it was resulted of word "bateria". So, the term "drums" replace the word "bateria" in sentence performing the automatic replacement activity. Next, the fragments are grouped again resulting in sentence "minha **drums** esta ruim" which is sent to machine translation to obtain the translation completed of sentence.

With the use of method, the homograph word "bateria" was translated correctly even it being contained in a disconnect sentence. Method considered through ontology the intent of user in conveying the idea of musical instrument. Therewith decreasing the ambiguity generated by term in short text translations. If the dialogue did not have the intelligence level employed by the method, the word "bateria" would probably be translated as "battery", an electronic component, since this is the most used by machine translation with statistical approach.

### 4.3. Example – Locution

This section illustrates the use of the tool to translate a sentence containing the homograph word accompanied by a locution. Portuguese Sentence "a guitarra estava ligada na bateria do carro." was translated to prove the addressing this problem. In this example was selected the ontology *Music Ontology* simulating a help desk about musical instrument.

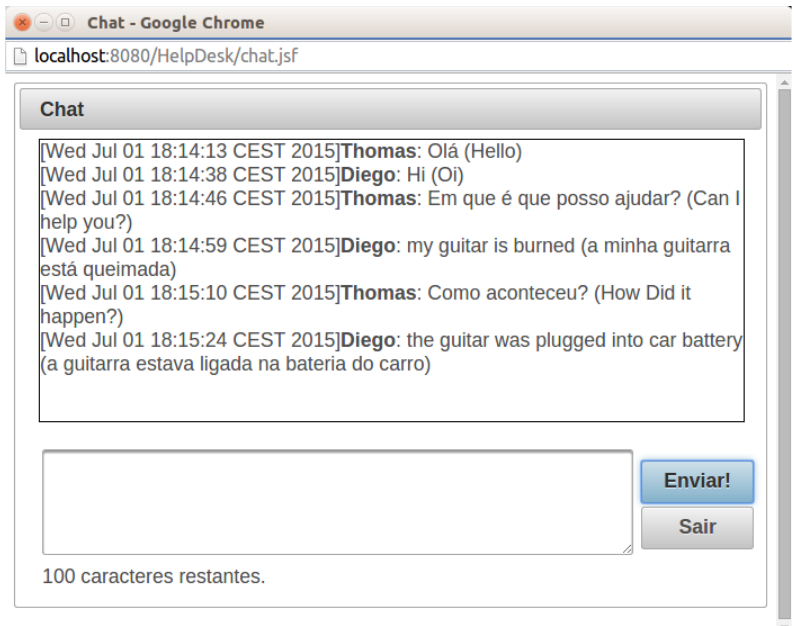

Figure 7. Locution - Music context

In Figure 6, the sender answered the message "a guitarra estava ligada na bateria do carro". Thus, the message goes through morphological analysis step. In process is identify that there is a possible homograph word. This term is word "bateria", but it is accompanied of a complement. Then the complement is noun locution being "do carro". Therefore the term "bateria" is not submitted to semantic analysis. Word is sent directly to translation step, it resulting in message "the guitar was plugged into car battery".

In translation is possible to notice that the locution "do carro" influenced directly in the translation of word. This influence is done when tha machine translation recovers the word "bateria" is part of car and it is not musical instrument as would be if the world is translated through ontology. This example presents the solution to locution and as Bing translation use statistical skill to translate sentences.

### 4.4. Dialogue Theory

In figure 7 shows an example using the method to address sub-contexts in dialogues. This example is based in dialogue theory. Theory is implemented through a temporary log. *Vehicle ontology* [32] was added in tool to provide the context. A sentence can be exemplify this technique is "I put the drums and the spark plugs at the trunk" as previously presented.

Through this sentence, the sender informs to receiver in target language that it put the drums and the spark plugs at the car. Therewith, the sentence goes through by morphological analysis process, reporting to method the momentary meaning of ambiguous term. These meanings are stored in temporary log. In sentence, the values "drums" to "bateria" and "sparkplugs" to "vela" are saved. When the message "Ok e a bateria estava completa?" is sent the translation of term "bateria" is retrieved from temporary log resulting by "ok, the drums was complete?". In case if this solution had not been implemented based in dialogue theory, the translation of term would be "battery" based in vehicle ontology.

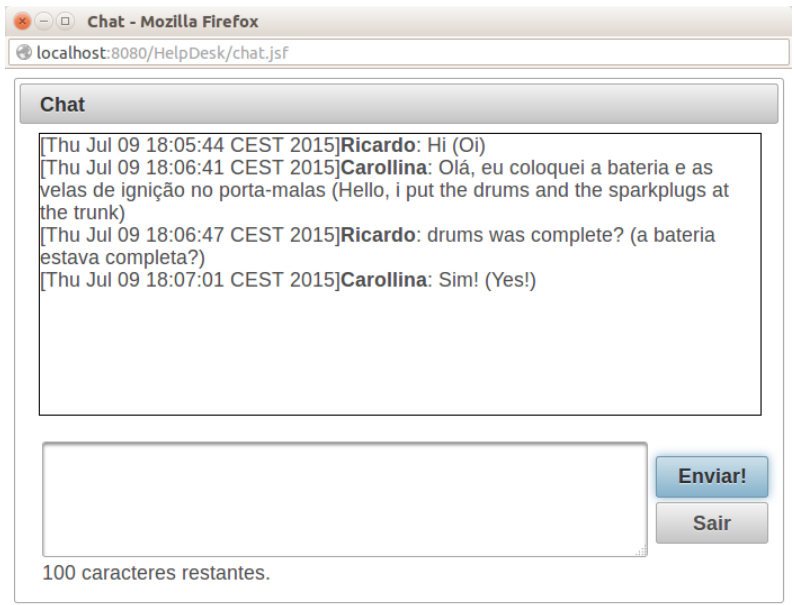

Figure 8. Dialogue theory - Vehicle context

### 4.5. Evaluation

The test were done of the follow way. First was add to dialogue only machine translation and checked separately the translation in dialogue. After was inserted the method to evaluate the disambiguation in dialogue. The method was checked together with different machine translations. To create the experiments was used semantic correctness as specific criteria talking into consideration the other criteria are corrected. As evaluation metrics was done in similar form that *Nyberg* [33]. In the test case were used 100 sentences each context.

Table 3. Evaluation with and without Judge Method.

| System/Word | Music | Electronic/ Computer | Vehicles | Sports | Financial |
|---|---|---|---|---|---|
| Bing Translator | 0.22 | 0.25 | 0.70 | 0.34 | 1.00 |
| Bing + Judge | 0.92 | 0.90 | 0.89 | 0.86 | 1.00 |
| Google Translator | 0.30 | 0.51 | 0.82 | 0.40 | 1.00 |
| Google + Judge | 0.86 | 0.74 | 0.86 | 0.91 | 1.00 |
| Worldlingo | 0.06 | 0.14 | 0.50 | 0.36 | 0.75 |
| Worldlingo + Judge | 0.67 | 0.70 | 0.77 | 0.52 | 0.93 |
| Gengo | 0.05 | 0.33 | 0.34 | 0.25 | 0.67 |
| Gengo + Judge | 0.24 | 0.72 | 0.56 | 0.49 | 0.85 |
| Systran | 0.13 | 0.23 | 0.53 | 0.60 | 0.67 |
| Systran + Judge | 0.70 | 0.70 | 0.78 | 0.88 | 0.94 |

With the test was possible to conclude that the pre-translation of homograph word only is effective with context a little bit used. For example, the word "banco" and "bolsa" in portuguese had 100% hits in financial context because yours translates like "bank" and "stock" is very used in statistical machine translation [34]. Nevertheless, the homograph word "vela" in sport context only effective with the method Judge when it supported the machine translation, because your translation "sail" is not very used at least the translation "candle" that appeared in many outputs. After experiments, the conclusion is necessary to define a pre-context before starting the chat, because the homographs word in short sentences is translated based on frequency table as showed before. Therefore, if the user is speaking about another subject the translation will result in error.

## 5. STATE-OF-THE-ART

As seen before there are many articles related with WSD (word sense disambiguation), but the focus is totally different, because this work presents that WSD using statistical approaches without knowledge semantic when applied to dialogue is not good. Before creating JUDGE method was considered these articles as inspiration;

In *Seo Eugene et. al.* [35] presents a machine translation using ontology as syntactically as semantically. This work disambiguates the word using weight among property existing within ontology. When is found an ambiguous word, first is queried your part of speech and of yours neighbours. After the word is checked if it is a possible polysemy word based on ontology description. In the end is retrieved the weight of each word and your following words, with that is done the disambiguation based on their properties.

Second work that served as baseline was *Shi, C., & Wang, H* [36]. This work also uses machine translation based on ontology. However, it does not put weight among relationship property. The disambiguation is carried out with base in the classes like person, animal and things. When is found an ambiguous words is checked your part of speech, thus is recovered possible meanings, then based on next word is compared what is the classes used together with another. E.g. the word "take" in Portuguese there are many meanings, therefore is needed to know the next word to find out the right meaning. This work retrieve it through hierarchy concepts existing in the ontology created by them.

Both works were developed to a specific language as spoken by authors. First to Korean-English and second Chinese-English. Each language has grammar rules therefore for machine translation is totally based on ontology is required the development to a specific language.

International Journal of Artificial Intelligence & Applications (IJAIA) Vol. 6, No. 5, September 2015

Although these works had served as brainwave, the focus was different, and then was selected two papers that use semantic web to improve the machine translation in short sentences.

Gutierrez [10] proposes an improvement of statistical algorithms *PageRank + Frequency* [37] with semantics features. This work makes the multilingual terms disambiguation to be with application in its proper language or in a possible translation. In *Harriehausen-Muhlbauer* [11] implements a novel approach developing a semantic machine translation. This machine performs automatic translation of a sentence based in relationship semantics into ontologies.

In Gutierrez [10], the work introduces a novel approach unsupervised to sense disambiguation of multilingual words. Main goal of work is provide automatically the right mean of an ambiguous term in differences idioms in a specific context. System proposes to be able to do this induction because it has a huge corpora originated by *BabelNet* and it also counting with use statistical algorithm as main method to do disambiguation. *BabelNet* is semantic tool created with union of *Wikipedia + WordNet* [38].

In its architecture after being selected many candidates as possible meanings to ambiguous term through BabelNet, It is used statistic algorithm to do the disambiguation. Nevertheless, even using semantics concepts to get the right candidates the method used of statistical model to get a conclusion. Therewith the result generates an uncertain when is chose the meaning, letting with probability the responsibility of induction of the result.

Idea imposed for *Harriehausen-Muhlbauer* [11] is presented as the future of automatic machine translations. Machines need to do the translation with base in semantics relationship extracted of an ontology. it is necessary and very important to use ontologies as an intelligence layer in dictionary. Author praises the necessity of machine translation have knowledge semantic to disambiguate words. A statistical machine translation use algorithms mathematics to realize a translation. These machines based its translation in repository that it has probabilistic data about each term. Therewith it is not possible a machine of this approach makes a disambiguation terms with good quality and high success rate.

As the proper author said in your work, the simple translation obtained a good quality. However, the time that it took to machine get to do translation was high a lot. Therefore, it used this approach to do a translation more complex is unpractical and it still more to communication in real time. Work used same tool and same layer pattern to do concept extraction that this our approach propose, both its use *SKOS*. This pattern is seen as main way to achieve the translation goal with ontologies, because it is responsible for describe concepts and exemplify it in more of an idiom.

Method presented in this paper has as focus the disambiguation of homographs words in a dialogue. A theme that it related works does not approach in its researches. In a dialogue must define a context to it has translation more right. This affirmation take consideration the small exchange of messages in a chat. Therefore, if use the Gutierrez method in communicator real time the result would not be satisfactory because of the statistic algorithm used to disambiguation and by answer time. As it said to realize a translation simultaneous in a dialogue, the machine must know about context. It is not happen with the work of Gutierrez, at send a disconnect sentence the method would result as a statistical machine translation pattern having in seen the example provided by author.

At realize a supposition as would be an implementation of solution propose by *Harriehausen-Muhlbauer* [11] in a dialogue system, the critics establishes by proper author were analysed. Main of them is that to send a sentence to tool, it would not take a time capable to do the translation. Use of ontologies in machine translation is very satisfactory just as decision layer to be to

30

disambiguation or domain specification. It exclusive utilization to texts translation still is deficient for time aspect. Current approaches to text translation has high level processing, but it does not have good quality that would be found with ontologies support.

Only messages in direct order were used in the execution tool. These messages use the regular form of Portuguese language i.e. subject + verb + complement. Messages followed the standard structure of Portuguese language. So when messages were sent without this pattern for example it putting the subject in end of sentence, the method presented difficulty to induce automatically the right mean showing a wrong translation. In morphological activity is necessary to add a syntactical activity to put the messages in direct order or formal order.

There was a difficulty to find ontologies that treat of domains and that could serve the method requirements. Only two ontologies that can be used perfectly by method without necessary of extended them. *Music Ontology* to music domain and *Vehicle Sales Ontology* to vehicles domain both terrestrial as aquatics.

## 6. CONCLUSIONS AND FUTURE WORKS

Ambiguity is a problem that it is open by the scientific community, because it does not have significant results and to generate many sub-problems. The exchange of messages in a dialogue, in both forms like writing or spoken is realized with base on a subject. This subject is responsible for generating the meaning of words, where some terms can have more than one meaning. Proposed approach focused an ambiguity sub-problems that is homographs words with it just goal to decrease the significant ambiguity in textual dialogues.

A tool had been developed to prove the effectiveness of the proposed method. Tool was coupled as a superior layer at translation component. In developing of tool had been taken into account the independence idea of components. Components responsible for each step can be replaced without affecting the structure of method, since the new components implement the same functions.

With the method was possible to prove that to do the translation simultaneous of dialogue between two person of nationality distinct and that it does not speak the same language is necessary to assign a context. This context may be attributed clearly by ontologies. First tests were done in a help desk, because it is only addressed one subject. Therewith it becoming the insertion of ontology more easily to context definition. Method presented promising and may be extended using of the multiples ontologies.

Currently is only possible to include one ontology as pre-context for time in the method. Next goal is not only used ontologies file, but also using Linked Data e.g. DBpedia [39] and will be implemented NER (named entity recognition) based on AGDISTIS [40] in the method. The idea is provide more resources in dialogue where the user can acquire more knowledge. The main goal is the people can change subject in dialogue without necessity of exchange log messages manually in the repository. For it evolution happening will be necessary to develop a method that search ontologies of automatic way in the web. Some test were done with *LOV API* [41] but the output needed to be improved and also using machine learning techniques in temporary log to create one supervised model to each user of dialogue.

Proposed solution in this paper introduced a novel optical related with simultaneous translations in a dialogue systems. It reaffirming the choose necessity of a previous context to achieve the translation more suitable, due presence of ambiguous words. However, user may speak about other contexts and contexts. Therewith, future works will be based in development of user model considering real user to have an alleged definition about your subject favourites and topics more related with your receiver. This information will retrieve on social network of user.

Furthermore, the method was able to bring up novels questioning, amplifying the research horizons in an area explored a lot, but with divergent results or a little satisfactory. Registered continuity of opportunities encouraging the production of scientific and technological knowledge to explore the scope of linguistics and machine translation into dialogue systems.